\documentclass[runningheads]{llncs}

\usepackage{eccv}

\usepackage{eccvabbrv}

\usepackage{graphicx}
\usepackage{booktabs}
\usepackage{wrapfig}
\usepackage[accsupp]{axessibility}  % Improves PDF readability for those with disabilities.
\usepackage{multirow}
\usepackage{multicol}

\usepackage{hyperref}

\usepackage{orcidlink}

\begin{document}

\title{Group-Equivariant Poincaré Convolutional Networks} 


\author{Aiden Durrant\inst{1} \and
Rahul Baburajan\inst{2}\and
Georgios Leontidis\inst{2}}

\authorrunning{A.~Durrant et al.}

\institute{School of Computing Sciences, University of East Anglia, NR4 7TJ, Norwich, UK \email{aiden.durrant@uea.ac.uk}\\\and
Department of Physics and Technology, UiT The Arctic University of Norway, NO-9037, Tromsø, Norway
}

\maketitle

\begin{abstract}
  While recent methods like that of the Poincaré ResNet have demonstrated the ability to learning visual representations directly in hyperbolic space, their optimisation remains a challenge, primarily due to the parameter redundancy of learning distinct orientation filters. In addition, hyperbolic learning exhibits distinct computational overheads that limit their wide use, where efforts to improve their efficiency via optimisation have seen good success, there has been limited exploration into structural priors that enable stronger sample efficiency at training. To address this, we propose Equivariant Poincaré ResNets, combining hyperbolic geometry with discrete symmetry groups ($C_4$ and $D_4$). We identify critical roadblocks in applying Euclidean equivariance to hyperbolic space and propose geometrically safe tensor reshaping, left-regular permutations for hyperbolic group convolutions, and joint-orientation Poincaré Midpoint Batch normalisation. Empirical evaluations show that embedding equivariance significantly improves the sample efficiency during training which in-turn accelerates convergence while respecting the boundary constraints of the Poincaré ball and retaining spatial group equivariance.\\
  Code: \href{https://github.com/AidenDurrant/GroupEquivariantPoincareCNN}{https://github.com/AidenDurrant/GroupEquivariantPoincareCNN}
\end{abstract}

\section{Introduction}
\label{sec:intro}

Deep learning in hyperbolic space has shown impressive abilities for embedding hierarchical visual data with minimal distortion\cite{krioukov2010hyperbolic,atigh2022hyperbolic,mishra2026hyperbolic}. The recent introduction of Poincaré residual networks has pushed this boundary further, enabling the learning of visual representations entirely within the Poincaré ball from the pixel level \cite{van2023poincare}. However, optimising deep convolutional networks on Riemannian manifolds introduces major issues when attempting to learn spatial symmetries efficiently.

A major limitation and computational inefficiency of current hyperbolic networks is their inability to recognise spatial symmetries. An object rotated by 90 degrees is treated by the network as an entirely new hierarchical concept, forcing the optimiser to to learn redundant representations. In Euclidean space, one approach to solve this is through group-equivariant convolutional networks \cite{cohen2016group}. Such networks have been shown to be highly data-efficient \cite{bietti2021sample}, and thus we propose that the Hyperbolic networks computational overhead can be somewhat mitigated through use of symmetric priors. 

In this paper, we investigate how to embed $C_4$(rotations) and $D_4$ (rotations and reflections) spatial-group equivariance over Poincaré-valued feature fields directly into the Poincaré ResNet architecture. Incorporating discrete symmetries into hyperbolic feature maps requires adapting channel operations to remain compatible with hyperbolic geometry. We propose three primary contributions to enable the Equivariant Poincaré ResNets: (i) We adapt the $\beta$-scaling formulation of Shimizu et al.\cite{shimizu2020hyperbolic} for flattening and unflattening to allow orientation channels to be decoupled without exceeding the expected norms of the manifold. (ii) We present G-CNNs in the hyperbolic space, proposing hyperbolic lifting and group convolutions by projecting base tangent-space filters through discrete symmetry transformations and left-regular permutations. (iii) We enable the Poincaré Midpoint Batch normalisation to act jointly over group orientations to prevent the normalisation step from destroying the learned equivariance.

\section{Background and Related Work}
\subsubsection{Hyperbolic Computer Vision.}

Hyperbolic spaces, characterised by constant negative curvature, have an intrinsic capacity to embed complex, tree-like structures with arbitrarily low distortion \cite{krioukov2010hyperbolic,nickel2017poincare}. As such, representing hierarchical relations in continuous spaces is naturally well fit for hyperbolic spaces. Hyperbolic representation learning has performed exceptionally in modelling continuous hierarchies across natural language processing \cite{tifrea2018poincar,ganea2018hyperbolic,le2019inferring}, graph networks \cite{chami2019hyperbolic,naddeo2026hyperbolic}, and recommender systems \cite{zhang2026hmamba,Tran2018HYPERML}.

The computer vision community has begun leveraging hyperbolic geometry more readily, establishing that both image data and labels contain latent hierarchical structures. The introduction of hyperbolic image embeddings has thus improved uncertainty quantification and few-shot learning \cite{khrulkov2020hyperbolic}. Subsequent works have successfully scaled hyperbolic methods to metric learning, zero-shot recognition, video action recognition \cite{long2020searching}, and part-whole image segmentation \cite{vlasenkohyperbolic, mishra2026hyperbolic, atigh2022hyperbolic}.

However, a signification limitation still exists in that the majority of these methods restrict hyperbolic operations to the final embedding or classifier space. In this setting the visual representations are still learned through standard Euclidean networks, where the geometric benefits during early feature extraction are not exploited \cite{van2023poincare}. While the recent introduction of the Poincaré ResNet \cite{van2023poincare} has shown that hyperbolic visual learning directly from the pixel level can be achieved stably, optimising these deep networks continues to be challenging. Notably, they rely on computationally expensive operations, and like in the Euclidean setting they do not display structural inductive biases required to recognise spatial symmetries, which in turn lead to better sample efficiency. Relevant hyperbolic prerequisites are given in Appendix \ref{appendix:prereq}.

\subsubsection{Equivariant Deep Learning.}

Equivariant machine learning exploits structural symmetries in data to constrain models with known structural priors. The introduction of geometric inductive biases consistently leads to improved generalisation \cite{elesedy2021provably,  petrache2023approximation}, interpretability \cite{bogatskiy2024explainable}, and computational efficiency \cite{bietti2021sample}. Enforcing symmetries typically requires parameter-sharing mechanisms such as \cite{ravanbakhsh2017equivariance}, where they have explored designing model parameters to reflect equivariance over discrete group actions, alternatively, regularisation schemes have also been used to encourage weight sharing to induce symmetries \cite{shakerinava2024weight}. Extending this further, Equivariant MLPs (EMLPs) were proposed, decomposing arbitrary discrete and Lie matrix groups into finite sets of generators \cite{finzi2021practical}.\\\indent Designing equivariant architectures from scratch is a notoriously challenging task, therefore alternative pipelines aimed to learn group-agnostic networks have emerged through methods such as frame-averaging \cite{puny2021frame}, probabilistic symmetrisation \cite{kim2023learning}, or learnable canonicalisation \cite{kaba2023equivariance}. Additionally and one additional challenge, real-world data often exhibits imperfect symmetries, where research into symmetry breaking and relaxed equivariance has risen as a solution. Methods include modifying EMLP constraints to handle broken symmetries \cite{kaba2023symmetry} or introducing relaxed group convolutions that combine spatial invariance with a linear combination of basis filter banks \cite{elsayed2020revisiting, wang2022approximately}. \\\indent Euclidean convolutional neural networks still mainly rely on innate spatial translation equivariance. This results in a lack of robustness to other foundational geometric symmetries, forcing networks to use parameters learning redundant filters for objects in different orientations. Group Equivariant Convolutional Networks (G-CNNs) \cite{cohen2016group, kondor2018generalization, weiler2018learning} resolve this by formally extending equivariance to discrete symmetry groups, such as the cyclic group $C_4$ (discrete rotations) and the dihedral group $D_4$ (rotations and reflections). By transforming filters and routing feature maps through left-regular permutations, G-CNNs ensure that a spatial geometric transformation of the input produces a predictable shift in the feature space, thus preserving equivariance to the discrete $C_4$ and $D_4$ groups.\\\indent However, embedding strict group equivariance into hyperbolic neural networks is mostly unexplored and where investigated present distinct challenges. As highlighted by \cite{huang2024lorentz} symmetric representations in hyperbolic space often suffer from geometric distortions, which makes strict equivariance preservation critical, although existing architectures often break these symmetries through unconstrained projections to the tangent space. Furthermore, where translating Euclidean group convolutions directly to the Poincaré ball we are limited by the manifolds rigid boundaries. Standard Euclidean approaches for generating equivariant feature maps, such as tensor concatenation, decoupling, and reshaping, have the impact of artificially inflating gyrovector magnitudes which push the features outside the valid manifold. Therefore, establishing true equivariance in this domain requires a hyperbolic reconstruction of lifting layers, group convolutions, and batch normalisation to satisfy both discrete group symmetries and non-Euclidean geometric constraints.

% While theoretical frameworks have introduced structure-preserving operations like $\beta$-concatenation to stabilise basic hyperbolic feature aggregation \cite{shimizu2020hyperbolic}, these mechanisms have not yet been adapted to handle the discrete algebraic routing required for symmetry preservation. 

\section{Group-Equivariant Poincaré Convolutions}
A key opportunity in non-Euclidean learning is achieving high sample to offset the computational overhead that comes with hyperbolic operations \cite{van2023poincare}. Without explicit structural priors hyperbolic networks, as do Euclidean counterparts, redundantly learn semantic concepts across multiple orientations reducing convergence speed and generalisation ability. To address this we propose strict group equivariance within hyperbolic neural networks as a foundational proof of concept, focusing on the cyclic ($C_4$) and dihedral ($D_4$) groups. These groups represent the natural symmetries of the pixel grid which in-turn will enable geometric transformations without artefacts from interpolation \cite{cohen2016group}.

To formulate an Equivariant Poincaré block, we cannot apply standard Euclidean convolutions operations natively as the operations do not adhere to the geometric constraints of the hyperbolic space. Instead, we must reconstruct lifting convolutions, group convolutions, and batch normalisation techniques to simultaneously respect the algebraic structure of the symmetry group $G$ and the geometric constraints of the Poincaré manifold $\mathbb{B}_c^n$.

\subsection{Geometrically Safe Tensor Unflattening ($\beta$-Scaling)}\label{sec:unflat}

In Euclidean equivariant networks the formulation of group-structured feature maps relies heavily on unconstrained concatenation, reshaping, and decoupling of tensors (e.g., transforming a flattened vector of size $|G| \times C$ back into $|G|$ orientations of size $C$). If we were to apply these Euclidean reshaping operations directly to vectors in the hyperbolic Poincaré manifold, it would result in geometrically invalid operations as concatenation inflates the vector magnitude, pushing the vector beyond the valid boundary of the Poincaré manifold.

To overcome this limitation \cite{shimizu2020hyperbolic} introduced $\beta$-concatenation, which dynamically shrinks vectors based on a Beta function ratio during the concatenation phase preserving the expectation of the Poincaré norm. To enable equivariance it is necessary that we invert this scaling process, however, a true isometric unflattening operation does not exist for hyperbolic concatenation, we instead adapt their framework to propose a numerically stable inverse $\beta$-split for channel decoupling. Doing so allows us to separate spatial and group channels while preserving the features magnitudes.

We define a geometrically safe unflattening operation for a concatenated Poincaré output vector $y_{\text{flat}}$ by first mapping the concatenated output back to the locally flat tangent space at the origin via the logarithmic map:

$$t = \log_0^c(y_{\text{flat}})$$

Once projected into the Euclidean tangent space the geometric bounds are lifted, allowing us to safely decouple the dimensions from $(B, |G| \times C, H, W)$ to $(B, |G|, C, H, W)$.

Because the initial forward operation natively applied $\beta$-scaling to merge the channels the true lengths of these vectors remain distorted. Therefore, we restore the magnitude of the decoupled tangent vectors, $t_{\text{decoupled}}$, to accurately reflect their original semantic norm before spatial aggregation. 

For a channel dimension $C$ and a group size $|G|$, the structurally restored tangent tensor $t_{\text{group}}$ is computed as:

$$t_{\text{group}} = t_{\text{decoupled}} \cdot \frac{\beta_{C}}{\beta_{|G| \times C}}$$

where $\beta_n$ is computed via the log-gamma function \cite{shimizu2020hyperbolic} to prevent floating-point underflow.

Lastly, we transpose this tensor into the target layout $(B, C, |G|, H, W)$ and the structured vectors are then projected back onto the Poincaré ball using the exponential map.

$$y_{\text{group}} = \exp_0^c(t_{\text{group}})$$

While not a strict geometric isometry, this stable inverse $\beta$-split bounds the magnitude of the restoration to preserve the expectation of the norms, this in turn prevents vector norms violating the manifold boundary and ensures the discrete group orientations are explicitly separated.

\subsection{Poincaré Lifting Convolutions}
The initial layer of an equivariant network must perform a base-to-group mapping which is commonly formalised as a lifting convolution in Geometric Deep Learning, to project the input data from the standard spatial domain $\mathbb{Z}^2$ into the group-structured domain $G$. Applying rotations directly to Poincaré gyro-vectors requires complex and computationally expensive Möbius transformations, therefore for this proof-of-concept we construct our symmetric filter banks entirely within the tangent space.

Formally, the input to our network can be viewed as a feature mapping $f:\mathbb{Z}^2 \rightarrow \mathbb{B}^d_c$. Equivariance in this context is defined as spatial-group equivariance, where the group action $g \in G$ transforms the spatial domain and enforces the left-regular permutation of the orientation channels while the feature vectors themselves remain anchored as Poincaré-valued representations.

We define a $C_4$ or $D_4$ Poincaré lifting convolution that maps an input feature map $x \in \mathbb{B}_c^{C_{\text{in}}}$ to an output space $y \in \mathbb{B}_c^{|G| \times C_{\text{out}}}$. We begin by initialising a base spatial filter in the tangent space, $w \in T_0\mathbb{B}_c^n$. To build the required equivariant filter bank we apply the respective group transformations directly to $w$ in-line with \cite{cohen2016group}, where for the $D_4$ group ($|G| = 8$) we generate the eight transformations representing all distinct rotations and horizontal flips:

$$w_k = \text{rot90}(w, k) \quad \text{for } k \in \{0, 1, 2, 3\}$$

$$w_{k+4} = \text{rot90}(\text{flip}(w), k) \quad \text{for } k \in \{0, 1, 2, 3\}$$

These transformed filters in the tangent space are then stacked along the new group dimension producing a flat weight matrix $w_{\text{flat}}$. Here the parameter-sharing condition is strict, enforcing that the learnable weights exist solely as the single base filter $w \in T_0\mathbb{B}_c^d$. The filter banks for all $|G|$ orientations are generated via deterministic spatial actions, guaranteeing that the number of parameters remains entirely independent of the group size $|G|$. To perform the spatial convolution natively on the manifold we follow the aproach of \cite{van2023poincare} where we extract local spatial patches from the input feature map via a hyperbolic unfold operation. The flattened weight matrix serves as the parameters within a standard Poincaré fully connected layer $F_c$, which is applied to the unfolded patches $X_{kl}$:

$$h_{kl} = F_c(\beta\|X_{kl}; w_{\text{flat}}, r)$$

The resulting output is then processed through the geometrically safe unflattening (Section \ref{sec:unflat}) to produce the lifted, group-equivariant representation.

\subsection{Poincaré Group Convolutions}

Once the features have been lifted into the group $G$, all subsequent convolutional layers must map from $G \to G$. A standard hyperbolic convolution is insufficient for this task as it would aggregate spatial features while ignoring the cyclical and reflective relationships of the group channels. For equivariance to be held, if an input image is rotated the feature map must route its activations to the corresponding channel that matches that shifted orientation. One caveat is that dynamically routing hyperbolic activations on the fly is computationally expensive, so we instead encode this routing directly into the expanded tangent-space filters.

We define the Poincaré Group Convolution as an input feature map $x$ characterised by both spatial dimensions and $|G|$ orientation channels. Given that the input itself now has an explicit orientation axis, applying a transformation from the group $G$ to the filter requires that we not only rotate or flip the spatial dimensions but also permute the orientation channels inline with the left-regular representation of the group.

\subsubsection{Left-Regular Permutation.}
To encode this routing, we define a Cayley index matrix $\mathcal{I}_G$ that dictates the required permutation mapping for every transformation. For instance, rotating a $C_4$ feature map by $90^{\circ}$ cyclically shifts its orientation channels by one. Therefore, we pre-permute the orientation axis of the tangent-space weights in accordance with the inverse left-regular representation.

\subsubsection{Aligned Application.}
For each transformation $i \in \{1 \dots |G|\}$, the base filter $w_{\text{spatial}}$ is geometrically transformed, and its internal orientation axis permuted:

$$w_{\text{aligned}}^{(i)} = w_{\text{transform}}^{(i)}\left[:, :, \mathcal{I}_G[i], :, :\right]$$

\subsubsection{Safe Flattening.}
To computing the hyperbolic inner product the input tensor $x$ must first be safely flattened by merging the channel dimension and the group dimension $|G|$ via the standard $\beta$-concatenation algorithm\cite{shimizu2020hyperbolic} to prevent numerical instability and the propagation of NaN values. The unfolded flattened input is then multiplied by the stacked $w_{\text{aligned}}$ matrices within the tangent space. As the aligned filters were stacked, maintaining a $[C_{\text{out}}, |G|]$ layout, the resulting flat features are reshaped into the correct orientation channels during the magnitude restoration step. By linking the weights across all symmetries and utilising the left-regular permutation a single gradient update applied to the base filter $w$ which simultaneously updates the representation for each four or eight orientations.

\subsection{Joint-Orientation Midpoint Batch normalisation}

Residual networks rely extensively on batch normalisation to stabilise deep feature representations. While \cite{lou2020differentiating} proposed Fréchet mean normalisation for hyperbolic spaces, Poincaré Midpoint normalisation offers a computationally effective alternative \cite{van2023poincare, ungar2022gyrovector}. However, applying either of these normalisation techniques independently to the separate group channels of an equivariant network introduces a catastrophic failure mode. If the $0^\circ$ orientation detects a strong feature magnitude while the $90^\circ$ orientation detects nothing, independent normalisation will centre and scale both to possess identical means and variances ($\mu$ and $\sigma^2$). Importantly, statistics indexed by a fixed orientation channel do not commute with the left-regular permutation so per-orientation normalisation breaks equivariance.

To maintain strict equivariance the normalisation statistics must be computed jointly over the entire group dimension. We define the Joint-Orientation Poincaré Batch normalisation by first reshaping the input tensor $x \in \mathbb{B}_c^n$ from its structural shape of $(B, C, |G|, H, W)$ to a merged spatial layout of $(B, C, |G| \times H, W)$. As merging the discrete rotations into the spatial dimension rearranges the vectors without altering their underlying norms or positions we can compute a Poincaré midpoint over this expanded volume:

$$\mu_{\text{joint}} = \frac{1}{2} \otimes_c \frac{\sum_{i=1}^{|G| \times H} \lambda_{x_i}^c x_i}{\sum_{i=1}^{|G| \times H} (\lambda_{x_i}^c - 1)}$$

After computing this shared midpoint $\mu_{\text{joint}}$ and its corresponding variance $\sigma_{\text{joint}}^2$, the vectors are transported and scaled uniformly then the tensor is reshaped back to its explicit group structure. To guarantee that the relative magnitude differences between $C_4$ and $D_4$ orientations are preserved, we calculate the variance and Poincaré midpoint over the joint spatial and group dimensions, enforcing an explicit statistical condition that all orientation channels are shifted and scaled by the exact same scalar.

\section{Experiments}
% \subsection{Experimental Setup}
All implementations of Hyperbolic layers, optimisation, and functions are handled by the Hypll library \cite{van2023hypll}, which we extend using the same convention for our proposed layers. We evaluate our architectures primarily on the CIFAR-10 dataset \cite{krizhevsky2009learning} and utilise a Poincaré ResNet-20 backbone, where to maintain manageable computation graphs and memory constraints inherent to Riemannian optimisation, we employ narrow channel widths of (8, 16, 32). Following \cite{van2023poincare} all convolutional layers are initialised using an identity-based mapping to prevent vanishing signals over deep layers and all Poincaré models are trained using the Riemannian Adam optimiser, while Euclidean models use standard Adam with a fixed learning rate of $10^{-3}$ and a weight decay of $10^{-4}$. Unless otherwise specified, all Poincaré models utilise layer-wise learnable negative curvature initialised at c=0.1 and further comparisons against a fixed static curvature (c=0.1) are detailed in the ablations.

Standard visual representation learning relies on significant data augmentation to force networks to learn spatial invariance. As our Equivariant Poincaré ResNet applies strict structural equivariance by design, it inherently recognises spatial transformations without this reliance, therefore, we isolate and evaluate this architectural inductive bias by intentionally omitting all spatial data augmentations from our training pipeline. While this setup results in a performance discrepancy compared to baseline literature utilising standard augmentations \cite{van2023poincare}, it ensures our evaluation provides a honest assessment of the networks geometric capabilities and sample efficiency.

\begin{wraptable}{r}{0.45\textwidth}
\vspace{-2em}
\caption{Top-1 accuracy (\%) on CIFAR-10 for ResNet-20. Results are compared across Euclidean and Poincaré manifolds with $C_4$- and $D_4$-equivariant layers. All Poincaré models use layer-wise learnable curvature initialised at c=0.1.}
\vspace{1em}
    \label{tab:main_results_wrapped}
    \centering
    \scriptsize
    % \linewidth adapts to the 0.5\textwidth of the wrap environment
    \begin{tabular*}{\linewidth}{@{\extracolsep{\fill}} l l c @{}}
        \toprule
        \textbf{Manifold} & \textbf{Group} & \textbf{Top-1 \%} \\
        \midrule
        \multirow{3}{*}{Euclidean} 
        & Standard & 78.26 \\
        & $C_4$    & 86.08 \\
        & $D_4$    & 88.69 \\
        \midrule
        \multirow{3}{*}{Poincaré}  
        & Standard & 76.87 \\
        & $C_4$    & 86.23 \\
        & $D_4$    & \textbf{88.77} \\
        \bottomrule
    \end{tabular*}
    % \vspace{-1em}
\end{wraptable}
\subsection{Classification Performance}
The addition of group equivariance into the Poincaré convolutional network results in a substantial improvement in both convergence efficiency and classification performance. As detailed in Table \ref{tab:main_results_wrapped}, we benchmarked standard, $C_4$-equivariant, and $D_4$-equivariant layers across both Euclidean and Poincaré manifolds where the baseline Poincaré ResNet-20 achieves a top-1 accuracy of 76.87\%. By applying the $D_4$-equivariant lifting and group convolutions, the Poincaré ResNet-20 reaches 88.77\% accuracy, a 11.9\% absolute improvement over the standard hyperbolic baseline, and outperforming the standard Euclidean ResNet-20 (78.26\%). These results validate that our proposed method adheres to the expected performance improvements established by the Euclidean baselines, and also marginally exceeds the Euclidean case.

\subsection{Rotational Robustness (RotMNIST)}
\begin{wraptable}{r}{0.55\textwidth}
\vspace{-3.5em}
    \caption{Test accuracy (\%) on Rotated MNIST. All Poincaré models use layer-wise learnable curvature initialised at c=0.1.}
    \label{tab:rotmnist_results}
    \centering
    \small
    \vspace{1em}
    \begin{tabular}{l c c}
        \toprule
        \textbf{Model} & \textbf{Group} & \textbf{Top-1 (\%)} \\
        \midrule
        Euclidean & Standard & 95.11 \\
        Euclidean & $C_4$ & 98.17 \\
        Euclidean & $D_4$ & 96.70 \\
        \midrule
        Poincaré & Standard & 93.66 \\
        Poincaré & $C_4$ & 97.92 \\
        Poincaré & $D_4$ & 96.15 \\
        \bottomrule
    \end{tabular}
    \vspace{-1em}
\end{wraptable}
% To explicitly validate robustness to spatial transformations, we evaluate our architecture on the MNIST-rotation benchmark introduced by Larochelle et al. \cite{larochelle2007empirical}. The dataset is constructed from MNIST digits by applying random rotations, with rotation angles sampled uniformly over the full angular range $[0,2\pi)$. Because the digits appear in unpredictable orientations, standard convolutional networks typically struggle to generalise without massive parameter inflation.

To explicitly validate robustness to spatial transformations, we evaluate our ResNet-20 backbone across both Euclidean and Poincaré manifolds on the MNIST-rotation benchmark introduced by Larochelle et al.~\cite{larochelle2007empirical}. The dataset consists of MNIST digits randomly rotated by angles sampled uniformly over $[0,2\pi)$, providing a standard benchmark for rotational robustness as well as equivariance.

As shown in Table \ref{tab:rotmnist_results}, standard spatial convolutions perform poorly on this task ($95.11\%$ for Euclidean, $93.66\%$ for Poincaré). Introducing discrete $C_4$ equivariance provides a structural prior, allowing the network to explicitly route rotated features rather than relearning them. The $C_4$-Equivariant Poincaré ResNet achieves a highly competitive $97.92\%$ accuracy, equalling its Euclidean $C_4$ counterpart. This confirms that the proposed hyperbolic operations preserve rotational symmetries under augmented spatial data. Notably, extending the symmetry group from $C_4$ to $D_4$ does not provide an additional benefit, and this finding is consistent with the structure of RotMNIST, where the benchmark introduces rotations, but not reflections as the transformation. Therefore, the additional reflection equivariance from $D_4$ presents an additional unnecessary inductive bias for this task, whereas $C_4$ more closely corresponds to the underlying transformation structure of the data.

\begin{figure}[htbp]
    \centering
    
    % ==========================================
    % LEFT FIGURE: Data Efficiency
    % ==========================================
    \begin{minipage}[t]{0.48\textwidth}
        \centering
        \includegraphics[width=\linewidth]{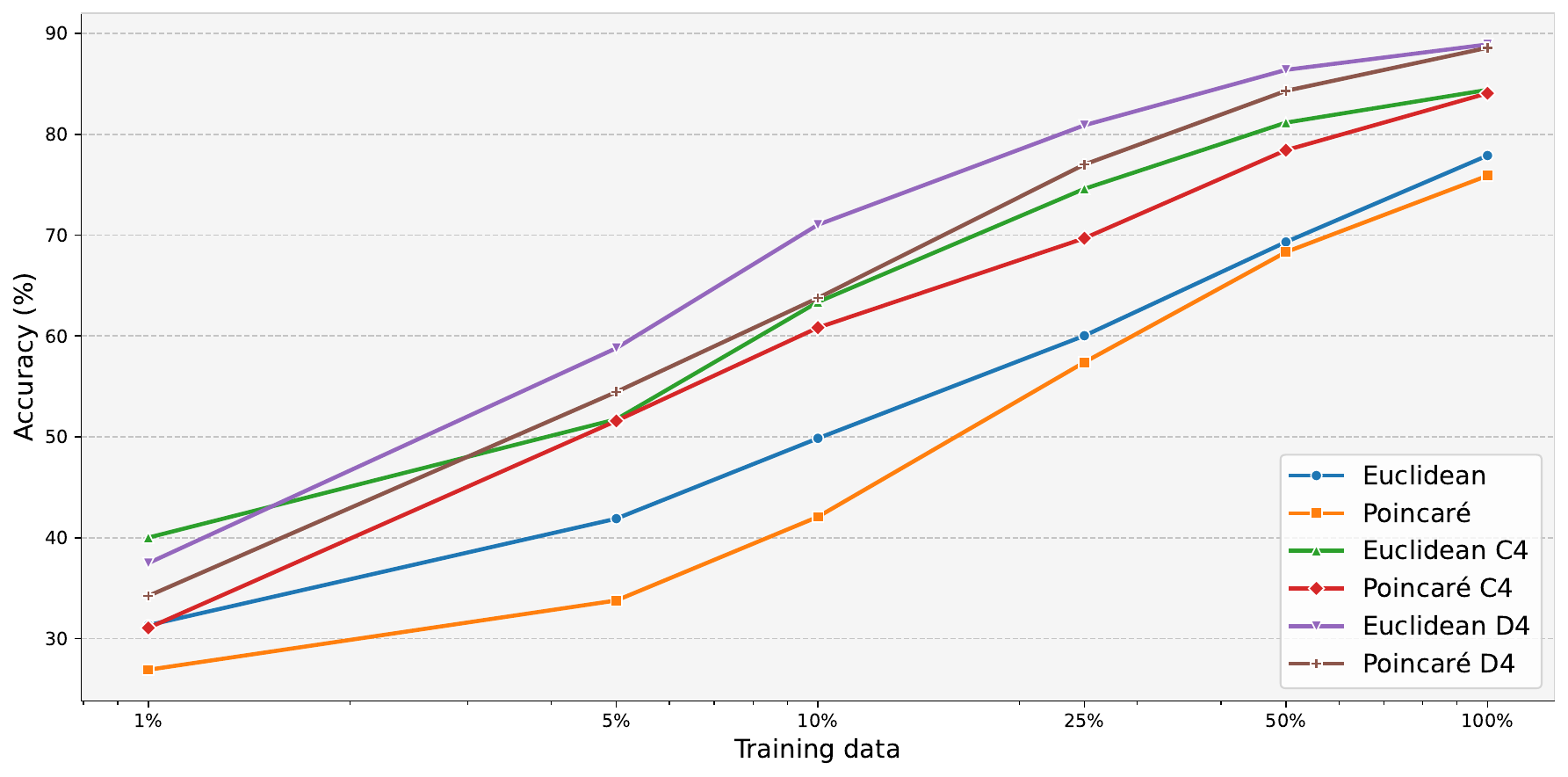}
        \caption{\textbf{Data-Efficiency Curve.} Top-1 test accuracy evaluated across discrete fractions of the training dataset. }
        \label{fig:sample_efficiency}
    \end{minipage}\hfill
    % ==========================================
    % RIGHT FIGURE: [Insert Your Second Figure]
    % ==========================================
    \begin{minipage}[t]{0.48\textwidth}
        \centering
        % Replace 'example-image' with your actual image path
        \includegraphics[width=\linewidth]{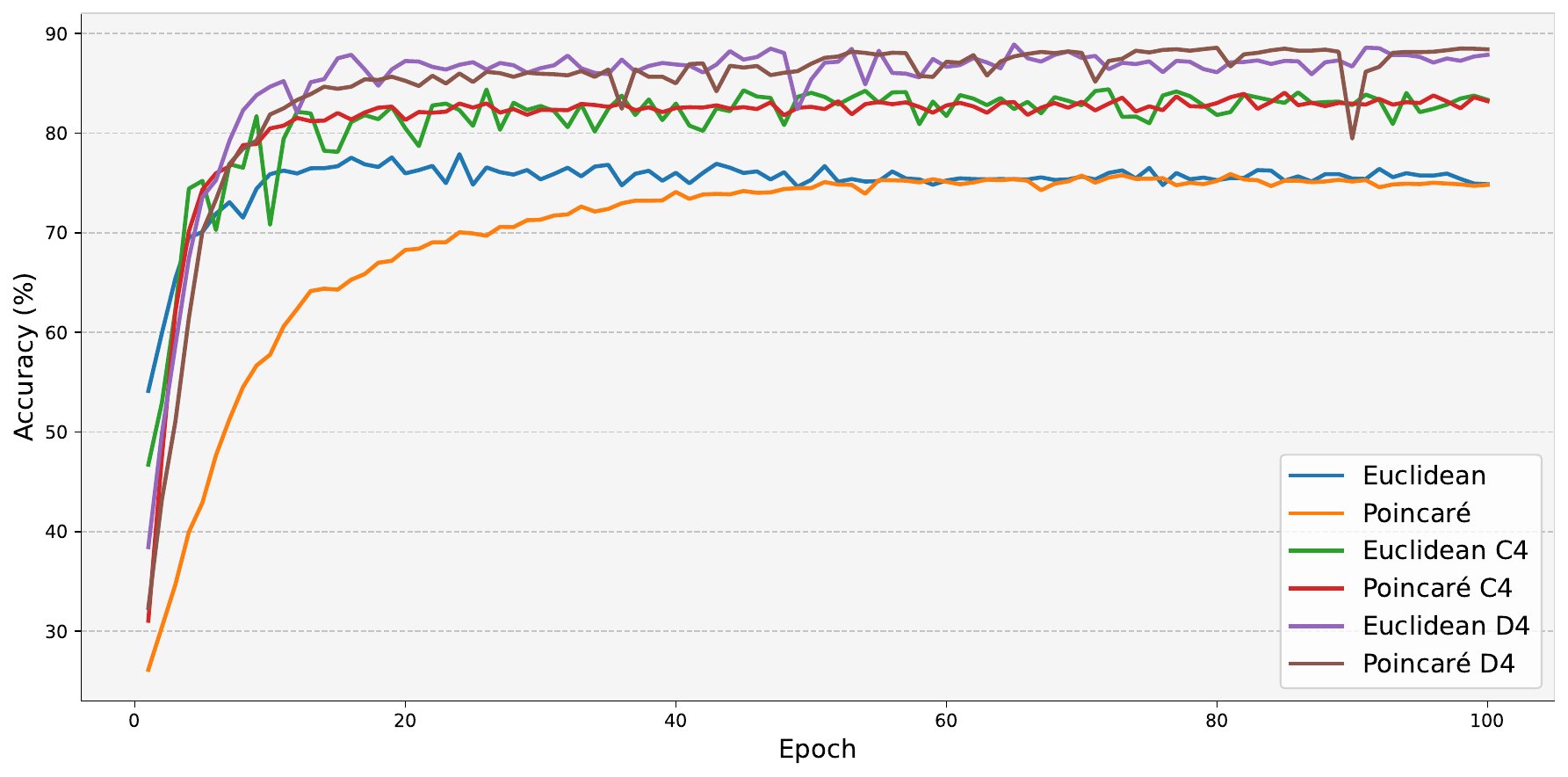} 
        \caption{\textbf{Convergence Trajectories.} Top-1 test accuracies evaluated during training for standard and equivariant Poincaré models. }
        \label{fig:convergence}
    \end{minipage}
    
\end{figure}

\subsection{Sample Efficiency and Convergence}
Embedding strict group equivariance significantly reduces the optimisation space, where without structural geometric priors standard hyperbolic networks redundantly learn identical concepts across varying orientations. To empirically validate that our architecture addresses these bottlenecks, we evaluate sample and convergence efficiency in a data-constrained setting.

To analyse sample efficiency we partition CIFAR-10 into class-balanced subsets representing 1\%, 5\%, 10\%, 25\%, 50\%, and 100\% of the total data and to isolate the inductive biases, we disable all spatial augmentations (e.g., cropping, rotations). The standard, $C_4$ and $D_4$ Euclidean and Poincaré variants of the ResNet-20 are then trained from scratch for 100 epochs per subset.

Figure \ref{fig:sample_efficiency} presents the sample-efficiency curve, we observe from this that in extreme low-data regimes (1\% and 5\%) standard Poincaré architectures degrade severely (achieving only 26.92\% and 33.79\% accuracy). Conversely, the $D_4$-Equivariant Poincaré ResNet demonstrates stronger generalisation, maintaining 34.24\% and 54.46\% accuracy on 1\% and 5\% of the data respectively. Furthermore, this structural prior in-turn results in notable improvements in convergence efficiency (Figure \ref{fig:convergence}). The equivariant models converge to peak accuracy in fewer epochs than standard networks in both the Euclidean and Poincaré settings, we argue this helps mitigate some of the computational overhead associated with hyperbolic learning. 
% \newpage
\subsection{Out-of-Distribution Detection}
\begin{wraptable}{r}{0.55\textwidth}
\vspace{-3.5em}
    \caption{Out-of-distribution detection on CIFAR-10 against Places365, SVHN, and Textures (DTD). Results are reported for ResNet-20 architectures using the energy score.}
    \label{tab:ood_detection_cifar10}
    \vspace{1em}
    \centering
    \scriptsize
    \begin{tabular}{l ccc}
        \toprule
        \textbf{Method} & \textbf{FPR95}$\downarrow$ & \textbf{AUROC}$\uparrow$ & \textbf{AUPR}$\uparrow$ \\
        \midrule
        \multicolumn{4}{c}{\textit{Places-365}} \\
        \midrule
        Euclidean     &  74.00 & 78.32 & 94.27 \\
        Poincaré      &  75.60 & 80.70 & 95.14 \\
        Poincaré $C_4$&  75.32 & 81.09 & 95.28 \\
        Poincaré $D_4$& \textbf{75.15} & \textbf{81.52} & \textbf{95.48} \\
        \midrule
        \multicolumn{4}{c}{\textit{SVHN}} \\
        \midrule
        Euclidean     & 98.45 & 66.42 & 92.32 \\
        Poincaré      & 74.55 & 80.72 & 94.02 \\
        Poincaré $C_4$& 73.04 & 82.91 & 96.03 \\
        Poincaré $D_4$& \textbf{72.95} & \textbf{83.17} & \textbf{96.17} \\
        \midrule
        \multicolumn{4}{c}{\textit{Textures}} \\
        \midrule
        Euclidean     & 92.87 & 66.91 & 92.12 \\
        Poincaré      & 80.27 & 73.93 & 93.31 \\
        Poincaré $C_4$& \textbf{74.06} & 80.33 & 95.49 \\
        Poincaré $D_4$& 74.15 & \textbf{80.94} & \textbf{95.59} \\
        \bottomrule
    \end{tabular}
    \vspace{-2em}
\end{wraptable}
To evaluate model robustness, we benchmarked our architectures ability to detect out-of-distribution (OOD) samples following the setup described in \cite{van2023poincare}. Models trained on CIFAR-10 (in-distribution) were evaluated against three highly distinct OOD datasets: Places-365 \cite{zhou2017places}, SVHN\cite{netzer2011reading}, and Textures (DTD)\cite{cimpoi2014describing}. Following standard evaluation protocols, we utilised the energy score \cite{liu2020energy} to distinguish between distributions, reporting the False Positive Rate at 95\% True Positive Rate (FPR95), Area Under the Receiver Operating Characteristic curve (AUROC), and Area Under the Precision-Recall curve (AUPR).

Detailed in Table \ref{tab:ood_detection_cifar10}, standard hyperbolic networks maintain the strong OOD robustness compared to their Euclidean counterparts even under equivariant constraints. This validates that our proposed method maintains the direct benefit of the boundary-aware hierarchical structuring of the Poincaré ball while simultaneously improving representational performance. For instance, on the SVHN dataset, the $D_4$-Equivariant Poincaré ResNet achieves an AUROC of 83.17 compared to the standard Poincaré model of 80.72, confirming that enforcing strict geometric symmetries does not compromise the manifolds natural capacity for outlier rejection. While Poincaré $D_4$ achieves the highest AUROC and AUPR across all benchmarks, the FPR95 on Places-365 remains comparable across models ($\approx 74$--$75\%$).

\begin{figure}[htbp]
    \centering
    
    % ==========================================
    % ROW 1: Vertically Centered Images
    % ==========================================
    \begin{minipage}[c]{0.48\textwidth}
        \centering
        \includegraphics[width=0.85\linewidth]{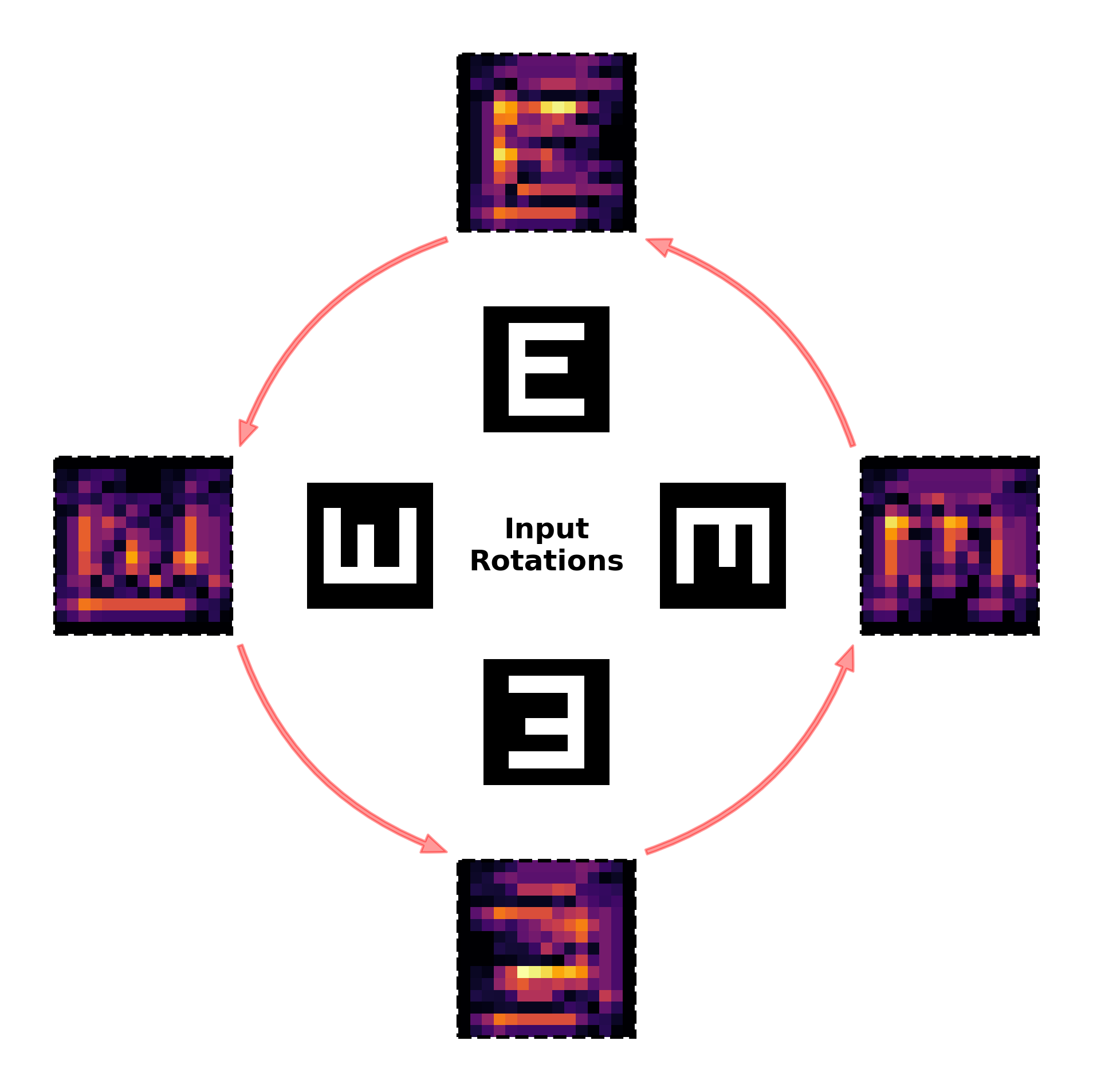}
    \end{minipage}\hfill
    \begin{minipage}[c]{0.48\textwidth}
        \centering
        \includegraphics[width=0.85\linewidth]{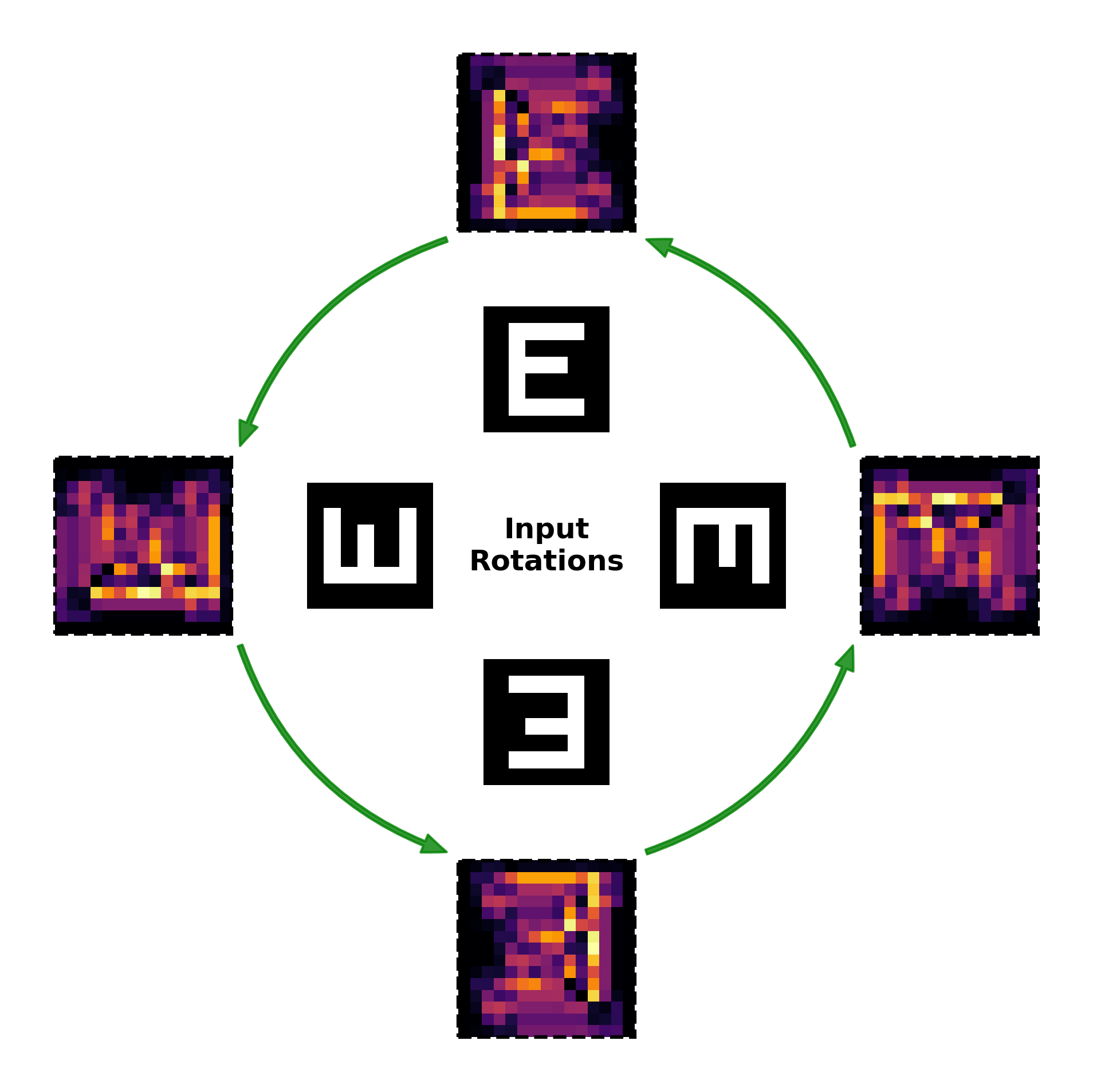}
    \end{minipage}
    
    \vspace{0.5em} % Small gap between images and their captions
    
    % ==========================================
    % ROW 2: Vertically Aligned Captions
    % ==========================================
    \begin{subfigure}[t]{0.48\textwidth}
        \caption{Standard convolution feature maps on the Poincaré manifold. The activations change unpredictably as the input rotates.}
        \label{fig:sub1}
    \end{subfigure}\hfill
    \begin{subfigure}[t]{0.48\textwidth}
        \caption{$C_4$ lifting convolution feature maps. The activations rotate spatially and shift cyclically across the four channels.}
        \label{fig:sub2}
    \end{subfigure}    
    \caption{\textbf{Visualizing $C_4$ equivariance.} Standard convolutions (a) fail to preserve the geometric relationship of the rotated inputs, whereas $C_4$-equivariant lifting layers (b) predictably rotate and permute the feature maps. Note, visualisations are projected onto the Euclidean plane. Features near the boundary appear distorted due to the Poincaré metric, yet the equivariant channel routing remains invariant to the distortion.}
    \label{fig:equivariance_visuals}
\end{figure}

\subsection{Equivariant Analysis}
To verify that our hyperbolic group convolutions preserve spatial symmetries without geometric degradation we empirically measure the relative equivariance error across our network layers at initialisation to ensure we are not learning the behaviour implicitly. While our architecture is equivariant by design, the projection of vectors between the Euclidean tangent space $T_0\mathbb{B}_c^n$ and the Poincaré manifold $\mathbb{B}_c^n$, combined with discrete tensor decoupling and floating-point approximations justifies the empirical validation. 

\subsubsection{The Relative Equivariance Error Metric.}

For a given layer $\Phi$, an input tensor $x \in \mathbb{B}_c^n$ and a spatial transformation $g \in G$, true equivariance states that transforming the input prior to the forward pass must produce the exact same result as applying the corresponding output transformation to the feature map.

Because the standard Euclidean $L_2$ distance is not a valid metric for comparing points directly on the coordinates of the Poincaré disk, we project the layer outputs back to the origin's tangent space via the logarithmic map $\log_0^c(\cdot)$. We define the tangent-space Relative Equivariance Error ($\Delta_{eq}$) as:

$$\Delta_{eq} = \frac{\|\log_0^c(\Phi(\rho_{\text{in}}(g)x)) - \log_0^c(\rho_{\text{out}}(g)\Phi(x))\|_2}{\|\log_0^c(\rho_{\text{out}}(g)\Phi(x))\|_2}$$

where $\|\cdot\|_2$ is the Frobenius norm, $\rho_{\text{in}}$ is the representation of the group action on the input space, and $\rho_{\text{out}}$ is the representation of the group action on the output feature space.

\subsubsection{Input and Output Group Representations.}
\begin{wraptable}{r}{0.5\textwidth}
\vspace{-2em}
\caption{Equivariance error ($\Delta_{eq}$) for various layer types across Euclidean and Poincaré manifolds.}
\vspace{1em}
    \label{tab:equivariance_results}
    \centering
    \scriptsize
    % resizebox automatically shrinks the font and table to fit the 50% width
    \resizebox{\linewidth}{!}{
        \begin{tabular}{lllll}
            \toprule
            Manifold & Type & Group & Transform & $\Delta_{eq}$ Error\\
            \midrule
            Euclidean & Conv & None & Rot 90 & \textcolor{red}{9.262756e-01}\\
            Euclidean & Lifting    & C4   & Rot 90 & 1.280077e-07\\
            Euclidean & Group      & C4   & Rot 90 & 2.967515e-07\\
            Euclidean & Lifting    & D4   & rot90  & 1.277800e-07\\
            Euclidean & Lifting    & D4   & fliph  & 1.162482e-07\\
            Euclidean & Group      & D4   & rot90  & 3.782480e-07\\
            Euclidean & Group      & D4   & fliph  & 4.742184e-07\\
            \midrule
            Poincaré  & Conv & None & Rot 90 & \textcolor{red}{1.350711e+00}\\
            Poincaré  & Lifting    & C4   & Rot 90 & 1.990071e-07\\
            Poincaré  & Group      & C4   & Rot 90 & 3.584960e-07\\
            Poincaré  & Lifting    & D4   & rot90  & 2.029714e-07\\
            Poincaré  & Lifting    & D4   & fliph  & 1.808202e-07\\
            Poincaré  & Group      & D4   & rot90  & 4.275676e-07\\
            Poincaré  & Group      & D4   & fliph  & 5.346042e-07\\
            \bottomrule
        \end{tabular}
    }
    \vspace{-1.5em}
\end{wraptable}
The definition of the transformation representations $\rho$ depends on the domain of the tensor. For the initial Lifting Convolution, the input lacks a group dimension, therefore, $\rho_{\text{in}}(g)$ acts solely as a geometric spatial transformation (e.g., rotating the $H$ and $W$ dimensions by $90^\circ$). For Group Convolutions where the input maps from $G \to G$ the features contain both spatial dimensions and a discrete orientation axis, $\rho_{\text{out}}(g)$ must simultaneously apply the spatial geometric transformation and permute the orientation channels according to the inverse left-regular representation. For the cyclic group $C_4$, this channel permutation is a simple cyclic shift, whereas, the dihedral group $D_4$ the permutation routes features across reflections and rotations employing a hardcoded index table mapping $g^{-1}h$. We therefore evaluate $\Delta_{eq}$ across both Euclidean ($\mathbb{E}^n$) and Poincaré ($\mathbb{B}_c^n$) manifolds for $90^\circ$ rotations ($C_4, D_4$) and horizontal reflections ($D_4$).

Table \ref{tab:equivariance_results} shows that standard hyperbolic convolutions fail the equivariance test ($\Delta_{eq} \approx 1.0$) demonstrating they are unable to structurally track rotational or reflective symmetries. Across all tested groups our proposed lifting and group convolutions yield a relative error bounded by 32-bit machine precision ($\Delta_{eq} \approx 10^{-7}$), confirming that our left-regular permutation logic and geometrically safe $\beta$-unscaling successfully preserve absolute non-commutative group symmetries.

\begin{table}[htbp]
\caption{Combinatorial ablation study of the Poincaré $D_4$ lifting layer on CIFAR-10. A \checkmark indicates the structural component is active, while a blank space indicates it was ablated. 10\% refers to the case where training failed with NaN. All Poincaré models utilise layer-wise learnable curvature initialised at c=0.1.}
    \label{tab:d4_ablations}
    \centering
    \scriptsize
    \setlength{\tabcolsep}{10pt} % Increases space between all columns
    \begin{tabular}{ccc cc}
        \toprule
        \multicolumn{3}{c}{\textbf{Model Components}} & \multirow{2}{*}{\textbf{Top-1 (\%)}} & \multirow{2}{*}{\textbf{$\Delta_{eq}$ Error}}\\
        \cmidrule(lr){1-3}
        $\beta$-Scaling & Joint-Oriented BN & Routing (Perm) & \\
        \midrule
        % 1. Baseline Model (No ablations)
        \checkmark & \checkmark & \checkmark & 88.77 & 4.428460e-07\\
        \midrule
        % Single Ablations (One feature removed)
                   & \checkmark & \checkmark & 85.65 & 4.456348e-07\\ % --ablate-beta
        \checkmark &            & \checkmark & 10.00 & 4.382712e-07\\ % --ablate-bn
        \checkmark & \checkmark &            & 81.85 & \textcolor{red}{1.426902e+00}\\ % --ablate-perm
        \midrule
        % Double Ablations (Two features removed)
                   &            & \checkmark & 10.00 & 4.373994e-07\\ % --ablate-beta --ablate-bn
                   & \checkmark &            & 79.72 & \textcolor{red}{1.411702e+00}\\ % --ablate-beta --ablate-perm
        \checkmark &            &            & 10.00 & \textcolor{red}{1.421677e+00}\\ % --ablate-bn --ablate-perm
        \midrule
        % Complete System Ablation (All features removed)
                   &            &            & 10.00 & \textcolor{red}{1.418058e+00}\\ % --ablate-beta --ablate-bn --ablate-perm
        \bottomrule
    \end{tabular}
    \vspace{-1em}
\end{table}

\subsection{Ablations}
To distinguish between components necessary for numerical stability and those required for algebraic equivariance, we evaluate both classification accuracy and the initialisation equivariance error ($\Delta_{eq}$) in Table \ref{tab:d4_ablations}.

The most significant ablationary factor is \textit{Joint-Orientation Batch Normalisation}, where standard Euclidean techniques compute statistics independently per channel, ablating our joint-orientation midpoint leads to catastrophic training failure (resulting in NaN values or 10.00\% random guessing). Here the independent scaling normalises each orientation on its own which entirely erases the relative magnitudes between the $D_4$ channels.

The \textit{Left-Regular Permutation Routing} is the primary mechanism responsible for equivariance, as when this routing is removed, the network aggregates spatial features without respecting cyclical relationships, as such, the structural equivariance is broken. This is demonstrated by the $\Delta_{eq}$ error spiking to $1.42$ at initialisation and classification accuracy degrades significantly to 81.85\%.

However, \textit{Geometrically Safe Magnitude Restoration} ($\beta$-scaling) functions as a numerical stabiliser rather than an equivariance necessity. When $\beta$-scaling is ablated, structural equivariance remains present ($\Delta_{eq} = 4.45 \times 10^{-7}$), but, downstream accuracy drops to 85.65\% due to standard unflattening inflating vector magnitudes.

Interestingly we observe a noteworthy sub-additive interaction when both $\beta$-scaling and permutation routing are ablated simultaneously. Under this setting, explicit equivariance is lost ($\Delta_{eq} \approx 1.41$), yet, this model still achieves 79.72\% accuracy, outperforming the standard non-equivariant Poincaré baseline (76.87\%). This indicates that Joint-Orientation Batch Normalisation provides a standalone regularising benefit and we conjecture that, by pooling normalisation statistics over the expanded volume (combining spatial and group axes), it stabilises hyperbolic feature variances independent of the equivariance constraints. We reserve further analyses for future work.

\subsubsection{Curvature.}
\begin{figure}[t]
    \centering
    
    % ==========================================
    % LEFT SIDE: FIGURE (Caption at the Bottom)
    % ==========================================
    \begin{minipage}[t]{0.5\textwidth}
        \vspace{0pt} % Forces the top baseline to match the right side
        \centering
        
        \includegraphics[width=\linewidth]{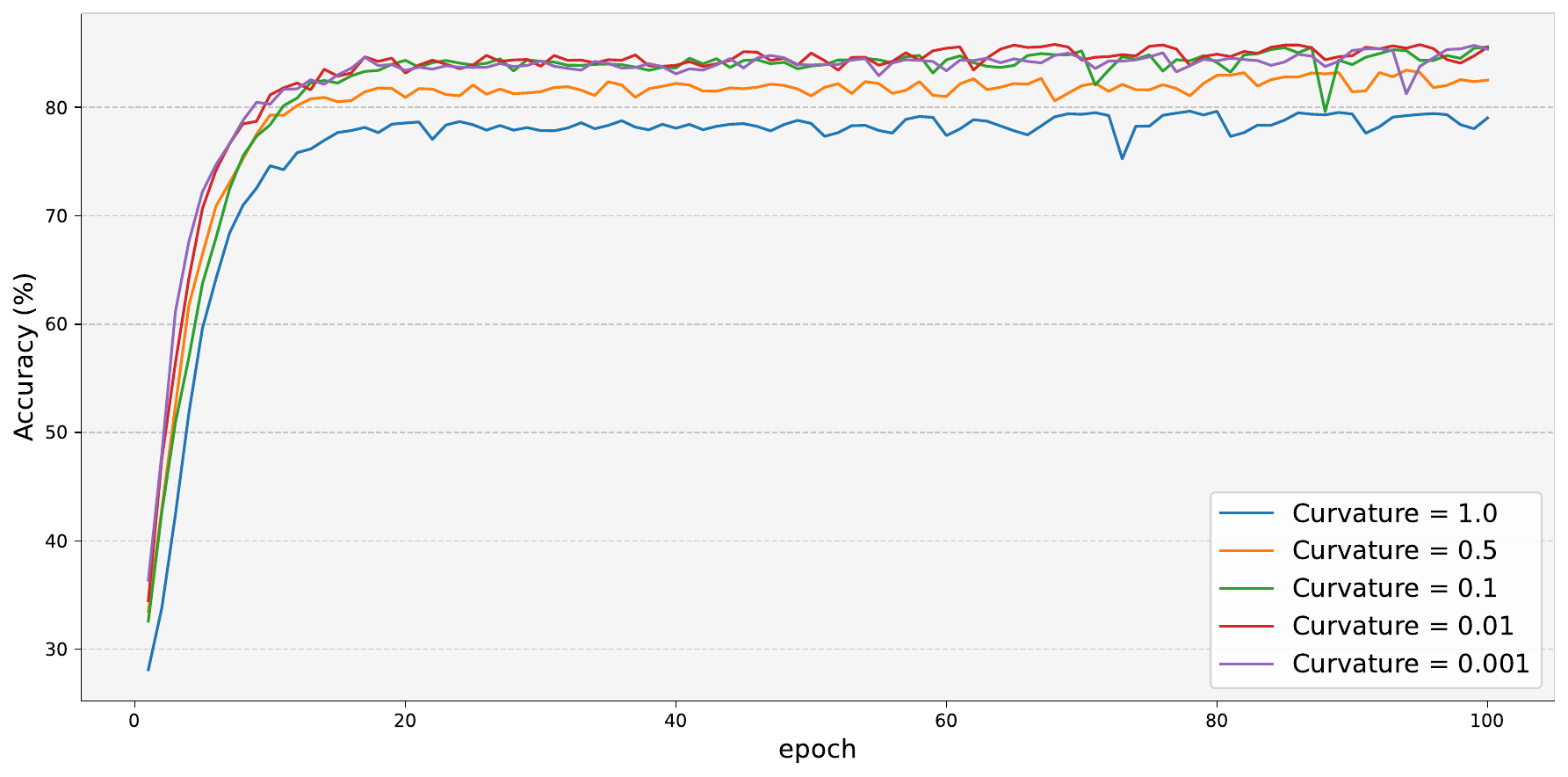}
        
        \caption{Effect of varying curvature magnitudes ($c \in \{0.001, 0.01, 0.1, 0.5, 1.0\}$) on CIFAR-10 performance within the Poincaré manifold.}
        \label{fig:curvature_sweep}
    \end{minipage}\hfill
    % ==========================================
    % RIGHT SIDE: TABLE (Caption at the Top)
    % ==========================================
    \begin{minipage}[t]{0.4\textwidth}
        \vspace{0pt} % Forces the top baseline to match the left side
        \centering
        
        \captionof{table}{Comparison of learnable vs. static curvature on the Poincaré $D_4$ model. Curvature is initialised to 0.1 when learnable, and set to 0.1 when static.}
        \label{tab:free_vs_fixed}
        
        \vspace{0.5em} % Small gap between the caption and the table
        
        \begin{tabular}{l c}
            \toprule
            \textbf{Curvature} & \textbf{Top-1 (\%)} \\
            \midrule
            Learnable & 88.77 \\
            Static    & 86.07 \\
            \bottomrule
        \end{tabular}
    \end{minipage}
    
\end{figure}

Finally we investigated the effect of the manifolds curvature constraint in both its initialised value and whether it is learnable. As shown in Figure \ref{fig:curvature_sweep} initialising with extremely tight curvatures (c=1.0) restrict the Euclidean volume of the Poincaré ball, making floating-point operations more unstable and resulting in suboptimal accuracy. However, our model remains generally robust to choice of curvature as shown in prior works \cite{van2023poincare}. Additionally, allowing the network to dynamically learn the curvature parameter per layer (Table \ref{tab:free_vs_fixed}) yields a top-1 accuracy of 88.77\%, outperforming a static curvature initialised at c=0.1 (86.07\%).

\subsection{Limitations.}
Our proposed equivariant Poincaré ResNets offer significant structural advantages, however, they inherit the computational overhead and floating-point instability at the manifold boundaries common to deep hyperbolic networks. This still requires narrow channel widths and rigorous numerical clipping which limits immediate scalability to high-resolution datasets like ImageNet. Specific to our equivariant formulation, the architecture is strictly bounded to discrete symmetry groups ($C_4$ and $D_4$) and cannot generalise to continuous transformations (e.g., $SO(2)$) without inducing significant interpolation errors across the parameter space. Additionally, our empirical validation is currently restricted to CIFAR-10, while this allows for controlled and rigorous testing of the architectural inductive biases, it leaves open the question of how these methods scale to more complex and larger-scale datasets. Future work would validate these findings across a broader range of benchmarks and domains with naturally occurring hierarchical structure where hyperbolic embeddings are expected to provide the greatest benefit.

\newpage
\section{Conclusion}

This work introduces Equivariant Poincaré Residual Networks, the first framework to embed discrete group equivariance ($C_4$ and $D_4$) directly into hyperbolic convolutional architectures. We developed a set of hyperbolic structure-preserving operations such as geometrically safe $\beta$-scaled tensor unflattening, left-regular permutation routing, and Joint-Orientation Poincaré Midpoint Batch Normalisation. These mechanisms ensure that discrete spatial symmetries are maintained without violating the geometric constraints of the Poincaré manifold, which we empirically validate by an equivariance error bounded at machine precision ($\Delta_{eq} \approx 10^{-7}$).

The empirical evaluations confirm that our approach does maintains strong task performance leading to an 88.77\% top-1 accuracy on CIFAR-10, which equates in a 11.9\% improvement over standard hyperbolic baselines. Furthermore, the embedded geometric priors significantly accelerate convergence and result in improved generalisation in low-data training settings, maintaining 54.46\% accuracy on just 5\% of the training data compared to 33.79\% for the standard Poincaré model under the same training conditions. By decoupling group orientations this work establishes a step toward highly efficient, geometrically grounded hyperbolic visual models. Future research will explore extending these principles to continuous symmetry groups, such as $SO(2)$, investigating the application in Hyperbolic Graph Neural Networks, and scaling to high-resolution, large-scale datasets.

\section*{Acknowledgements}
This work was funded, in part, by RCN (the Research Council of Norway) through Visual Intelligence, Centre for Research-based Innovation (grant no. 309439). The computations were performed, in part, on resources provided by Sigma2—the National Infrastructure for High-Performance Computing and Data Storage in Norway (Project NN8106K). In addition, part of the training and evaluation was carried out on the High-Performance Computing Cluster supported by the Research and Specialist Computing Support service at the University of East Anglia and in part by the AMD University Program.

\bibliographystyle{splncs04}
\bibliography{main}

\newpage
\appendix

\section{Prerequisites}
\subsection{Hyperbolic Learning: The Poincar\'e Ball Model}
\label{appendix:prereq}

Hyperbolic space $\mathbb{B}^n$ is the unique simply connected $n$-dimensional Riemannian manifold of constant negative curvature, where curvatures measure the deviation from flat Euclidean geometry. The constant negative curvature of the hyperbolic space, although analogous to the Euclidean sphere, presents some significant differences in geometric properties. As such hyperbolic space cannot be isometrically embedded into Euclidean space, yet there exist a number of conformal models of hyperbolic geometry \cite{cannon1997hyperbolic} employing hyperbolic metrics providing a subset of Euclidean space. In this work, we employ the Poincar\'e ball model for hyperbolic geometry given its wide adoption in computer vision and unique properties ideal for embedding between Euclidean and hyperbolic representations. The Poincar\'e ball model $(\mathbb{B}_{c}^{n}, g^{\mathbb{B}_c})$ is defined by the manifold $\mathbb{B}_{c}^{n}=\{x\in \mathbb{R}^n : c \|x\|^2 <1\}$ with the Riemannian metric 

\begin{equation}
    g^{\mathbb{B}_c} = (\lambda^c_x)^2 g^E = \left(\frac{2}{1-c\|x\|^2}\right)^2 \mathbb{I}^n
\end{equation}
where $g^E=\mathbb{I}^n$ is the Euclidean metric tensor and $\lambda^c_x=\frac{2}{1-c\|x\|^2}$ is the conformal factor with $c$, a hyperparameter, controlling the curvature and radius of the ball. The conformal factor scales the local distances which approach infinity near the boundary of the ball, providing the unique property of space expansion. Such space expansion of hyperbolic spaces makes them continuous analogues of trees, given volumes of an object with diameter $r$ scale exponentially with $r$. Thus, when referring to a tree with branching factor $k$, there are $\mathcal{O}(k^l)$ nodes at level $l$, where $l$ serves as a discrete analogue of the radius. This is the fundamental property which the advocating work \cite{ganea2018hyperbolic, khrulkov2020hyperbolic, yan2021unsupervised} and ours takes advantage of, allowing for the efficient embedding of natural hierarchies \cite{sarkar2011low}. 

Our approach employs encoders that operate in Euclidean space, and as such, we need to define a bijection from Euclidean embeddings of the encoder to the Poincar\'e ball of hyperbolic space. To achieve this we apply an exponential map $\exp^c_{v}(x): \mathbb{R}^n \rightarrow \mathbb{B}_{c}^{n}$ on Euclidean vector $x$ with some fixed base point $v \in \mathbb{B}_{c}^{n}$ which we set to be the origin, simplifying the exponential map and measures of distance which will be defined later. The exponential map is as follows,
\begin{equation}
    \exp^c_{v}(x) = v{\oplus}_{c} \left( \tanh \left( \sqrt{c}\frac{\lambda^c_v \|x\|}{2} \right) \frac{x}{\sqrt{c}\|x\|} \right)
    \label{eq:exp-map}
\end{equation}
with its inverse logarithm map given by
\begin{equation}
    \log^c_{v}(x) = \frac{2}{\sqrt{c}\lambda^c_v}\text{arctanh} \left( \sqrt{c}\|-v{\oplus}_{c}x\|\right)\frac{-v{\oplus}_{c}x}{\|-v{\oplus}_{c}x\|}
    \label{eq:log-map}
\end{equation}
Given the change in geometry, hyperbolic spaces do not allow for standard vector space operations, as such we employ gyrovector formalism for standard operations such as addition, subtraction, multiplication \cite{ungar2008gyrovector,ganea2018hyperbolic}. Therefore, from Eq. \ref{eq:exp-map}, ${\oplus}_{c}$ is defined as the gyrovector or M\"obius addition of a pair of points $x,y \in \mathbb{B}_{c}^{n}$ 
\begin{equation}
    v{\oplus}_{c}w =\frac{(1+2c\langle v,w \rangle +c\|w\|^2)v+(1-c\|v\|^2)w}{1+2c\langle v,w \rangle + c^2\|v\|^2\|w\|^2}
\end{equation} 
Leading from gyrovector formalism is the notion of distance, vital for self-supervised losses where typically the Euclidean cosine similarity and distance, are employed \cite{richemond2020byol,chen2020simple,bardes2021vicreg}. On the Poincar\'e ball of hyperbolic space we define the distance between $x,y \in \mathbb{B}_{c}^{n}$ as follows:
\begin{equation}
    \text{dist}_{\mathbb{B}}(x,y) = \frac{2}{\sqrt{c}}\text{arctanh}\left( \sqrt{c}\|-x\oplus_c y\|\right)
    \label{eq:dist}
\end{equation}
which with $c=1$ recovers the geodesic, %which with $c=1$ the geodesic is recovered, 
a vital concept given cosine similarity is analogous to sphere geodesic distance, whereas $c \rightarrow 0$ produces the Euclidean distance. 
\end{document}